\documentclass[lettersize,journal]{IEEEtran}
\usepackage{amsmath,amsfonts}
\usepackage{algorithmic}
\usepackage{algorithm}
\usepackage{array}
\usepackage[caption=false,font=normalsize,labelfont=sf,textfont=sf]{subfig}
\usepackage{textcomp}
\usepackage{stfloats}
\usepackage{url}
\usepackage{verbatim}
\usepackage{graphicx}
\usepackage{cite}
\usepackage{multirow}
\usepackage{booktabs}
\usepackage{makecell}
\usepackage{bbm}

\begin{document}

\newcommand{\sysname}{SARF}

\title{\sysname{}: Aliasing Relation Assisted Self-Supervised Learning for Few-shot Relation Reasoning}

\author{Lingyuan Meng$^{\ast}$,
        Ke Liang$^{\ast}$,
        Bin Xiao,
        Sihang Zhou,
        Yue Liu,
        Meng Liu, \\
        Xihong Yang,
        Xinwang Liu$^{\dag}$, ~\IEEEmembership{Senior~Member,~IEEE},

\IEEEcompsocitemizethanks{\IEEEcompsocthanksitem $^{\ast}$ Equal contribution.
\IEEEcompsocthanksitem $^{\dag}$ Corresponding Author.
\IEEEcompsocthanksitem Lingyuan Meng, Ke Liang, Xinwang Liu, Yue Liu, Meng Liu and Xihong Yang are with the School of Computer, National University of Defense Technology, Changsha, 410073, China. E-mail: {xinwangliu@nudt.edu.cn}.
\IEEEcompsocthanksitem Sihang Zhou is with the College of Intelligence Science and Technology, National University of Defense Technology, Changsha, 410073, China.
\IEEEcompsocthanksitem Bin Xiao is with the Chongqing University of Posts and Telecommunications, School of computer science, Chongqing, 400065, China.
}
\thanks{This work has been submitted to the IEEE for possible publication. Copyright may be transferred without notice, after which this version may no longer be accessible.}}



\maketitle


%












\begin{abstract}
Few-shot relation reasoning on knowledge graphs (FS-KGR) aims to infer long-tail data-poor relations, which has drawn increasing attention these years due to its practicalities. The pre-training of previous methods needs to manually construct the meta-relation set, leading to numerous labor costs. Self-supervised learning (SSL) is treated as a solution to tackle the issue, but still at an early stage for FS-KGR task. Moreover, most of the existing methods ignore leveraging the beneficial information from aliasing relations (AR), \textit{i.e.,} data-rich relations with similar contextual semantics to the target data-poor relation. Therefore, we proposed a novel \underline{S}elf-Supervised Learning model by leveraging \underline{A}liasing \underline{R}elations to assist \underline{F}S-KGR, termed \sysname{}. Concretely, four main components are designed in our model, \textit{i.e.,} SSL reasoning module, AR-assisted mechanism, fusion module, and scoring function. We first generate the representation of the co-occurrence patterns in a generative manner. Meanwhile, the representations of aliasing relations are learned to enhance reasoning in the AR-assist mechanism. Besides, multiple strategies \textit{i.e.,} simple summation and learnable fusion, are offered for representation fusion. Finally, the generated representation is used for scoring. Extensive experiments on three few-shot benchmarks demonstrate that \sysname{} achieves state-of-the-art performance compared with other methods in most cases. 
\end{abstract}


\section{Introduction}
\begin{figure}
    \centering
     \setlength{\abovecaptionskip}{-0.02cm}
    \includegraphics[width=0.49\textwidth]{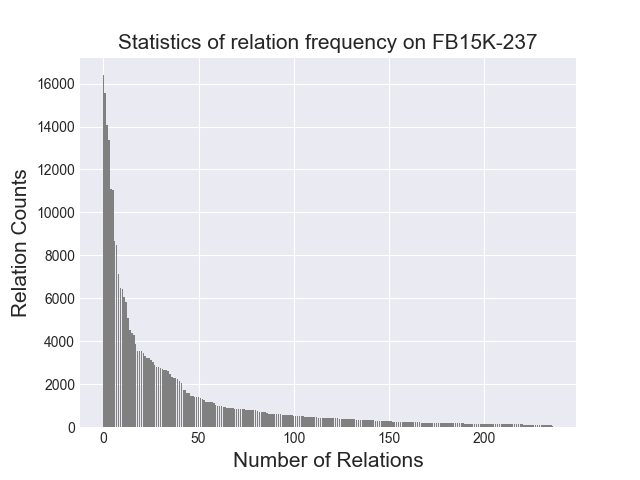}
    \caption{The statistics of relation frequencies of FB15K-237. FB15K-237 contains 237 relations and 310116 triplets. There are a large amount of long-tail relations with only few instances.}
    \label{fig1}
\end{figure}

\IEEEPARstart{K}{nowledge} graphs (KGs) organize massive multi-relational knowledge in a directed graph structure. They play an important role in many knowledge-driven downstream tasks \cite{oursurvey}, including information extraction \cite{dialog1,dialog2}, program analysis\cite{ABSLearn}, question answering\cite{QA1,QA2}, medical diagnosis \cite{medical} etc. However, incompleteness issues can be commonly found in most of the real world KGs, such as Freebase \cite{freebase}, NELL \cite{nell}, and WikiData \cite{wikidata}. To address the problem, it is essential to conduct missing fact prediction, i.e., knowledge graph reasoning (KGR) over these heterogeneous data structures. Over the past decades, a large portion of powerful approaches has achieved great success on KGR task \cite{grail,rgcn,NBF}. In particular, most methods assume the presence of a substantial number of instances for both entities and relations. However, an observation reveals that few-shot relations are commonly found in knowledge graphs (KGs) due to certain inherent properties of KGs. The first property is a large amount of long-tail data-poor relations \cite{gmatching} in real-world KG. As shown in figure\ref{fig1}, the FB15K-237 has about 40\% of relations corresponding to only 0.1\% of the total number of instances. Besides, KGs are updated continuously over time in many practical scenarios such as recommendation systems, computational medicine, and social media, which inevitably lead to the emergence of new relations with poor data. Existing KGR models cannot be well learned over long-tail data-poor relations. Therefore, it is necessary to infer relations with a few instances.

In past years, various few-shot KG reasoning (FS-KGR) approaches have been proposed over the meta-learning paradigm, which has achieved satisfactory performance. GMatching\cite{gmatching} and FSRL\cite{fsrl} aims to learn a differentiable metric that ensures positive query triplets are in close proximity to the embeddings of triplets in the support set, while negative triplets are far away from them. Att-FMetric\cite{attfmetric} predict long-tail relations based on attention neighborhood aggregation and path encoding. Meta-R\cite{metar} try to optimize the task learner with only a few instances by training a meta-learner. 

\begin{figure}
    \centering
    \includegraphics[width=0.47\textwidth]{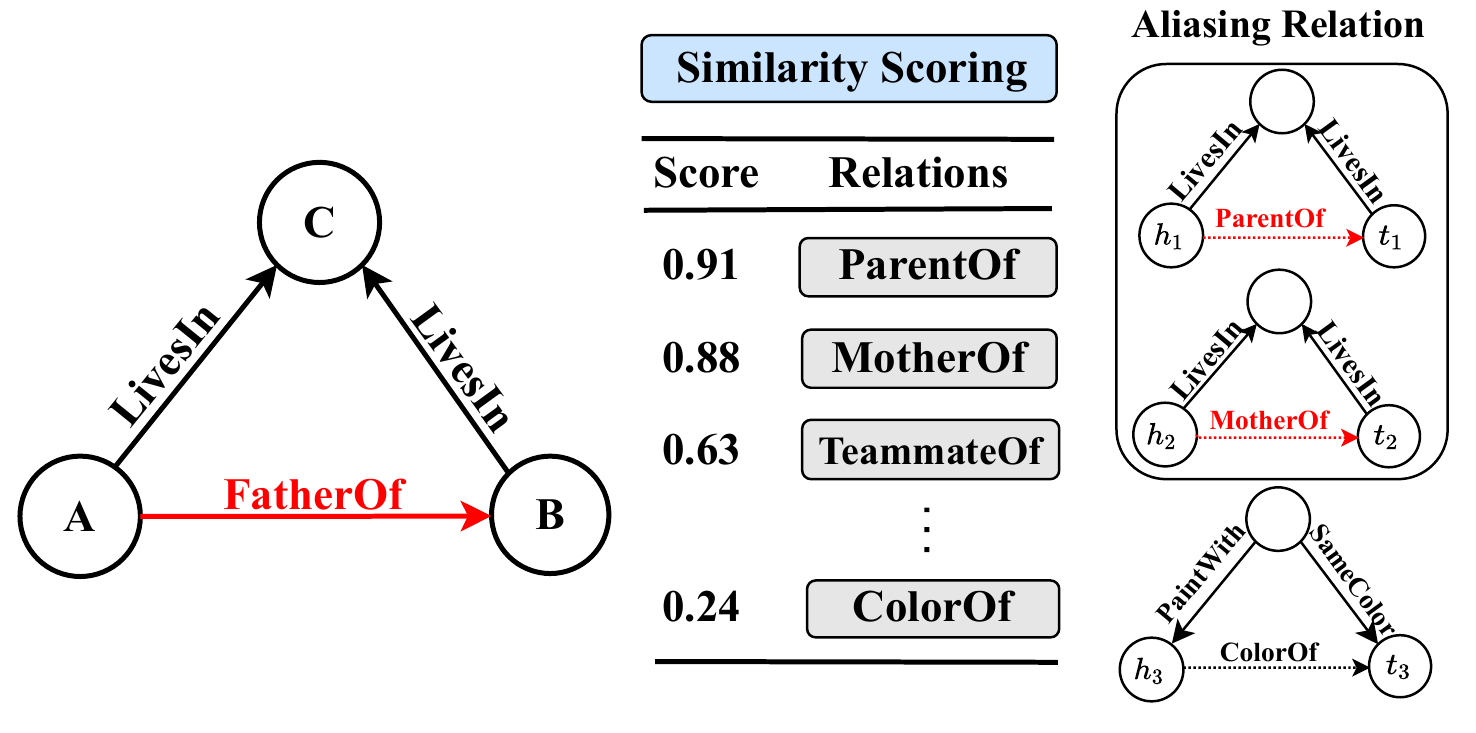}
    \caption{Illustration of aliasing relations. Aliasing relations ("ParentOf" and "MotherOf" in right figure) are selected from background KG over semantically similarity comparison with the target relation ("FatherOf" in left figure). }
    \label{intro}
\end{figure}

Although achieved competitive performance, drawbacks still exist in these methods. First, they depend heavily on the manual-created meta-training tasks towards specific relations. Concretely, the model is trained on numerous meta-training tasks during the training process, which leads to poor robustness if the meta-training set is not well-designed. Additionally, creating an excessive number of meta-training tasks from the background knowledge graph (KG) could result in the removal of a substantial number of edges, leading the KG to become sparse due to the limited number of relations. Such drawbacks limit the robustness and expressive ability of the model. 

Recently, self-supervised learning (SSL) techniques have been regarded as a more simple and robust paradigm to cast these problems in other fields, \textit{e.g.}, computer vision, by sufficiently digging out potential structure information underlying the data in a generative manner without introducing more meta-tasks \cite{lm1}. As our best known, only a few works, especially CSR \cite{csr}, try to extend the idea to FS-KGR tasks, leaving us gaps to fill. Moreover, existing SSL-based FS-KGR approaches depend heavily on subgraph structure, ignoring the logic patterns of data-rich relations which are useful to our task, especially for aliasing relations (AR), \textit{i.e.}, those relations that are semantically similar to the target relation. For example in figure\ref{intro}, the relation $ParentOf$ and $MotherOf$ are more semantically similar to the target relation $FatherOf$ than $ColorOf$, which is defined as ARs. Such ARs usually hold similar logic patterns to the target relation which are beneficial for reference. Intuitionally, the logical patterns mainly represent subgraph structure information. For example in figure \ref{intro}, the pattern around triplet $(h_1, ParentOf, t_1)$ and triplet $(h_2, MotherOf, t_2)$ are same as the $(A, FatherOf, B)$. Therefore, leveraging ARs to assist reasoning is an effective way to improve few-shot relation reasoning tasks.

To this end, we propose a novel SSL few-shot relation reasoning model, termed SARF, by leveraging \underline{A}liasing \underline{R}elations to assist \underline{F}ew-shot KG reasoning. Concretely, we divide our model into four main components, \textit{i.e.}, SSL reasoning module, AR-assisted mechanism, fusion module, and triplet scoring. We first extract co-occurrence patterns, \textit{i.e.}, the shared subgraph of all triplets in the support set, to represent the target relation by using a co-occurrence extractor. After that, we try to encode and reconstruct co-occurrence patterns in a generative SSL manner. Meanwhile, the AR-assisted mechanism is proposed in order to assist reasoning. Besides, multiple strategies, \textit{i.e.}, summation fusion strategy, and learnable fusion strategy, are proposed to fuse the representation of co-occurrence pattern and aliasing relations. Finally, the fused representation is used for triplet scoring. In this way, our SARF can achieve competitive expressive ability for few-shot relation reasoning with acceptable computational cost proven by the experiments. In conclusion, we summary four main contributions of this work as follows:
 
\begin{itemize}
    \item We proposed a novel self-supervised few-shot model for few-shot KG reasoning, SARF, which effectively leverages the information from aliasing relations. To the best of our knowledge, we are the first to make use of data-rich relations to assist the data-poor relation reasoning.

    \item  We design the aliasing relation-assisted mechanism. The aliasing relations are selected from data-rich relations according to semantic similarity. Besides, the aliasing relation-assisted mechanism can also be easily adopted by other FS-KGR models.

    \item We offered two strategies, \textit{i.e.}, summation fusion strategy and learnable fusion strategy, to fuse the representation of co-occurrence patterns with that of aliasing relations. 

    \item Comprehensive experiments conducted on three few-shot datasets substantiate  that our SARF outperforms existing other FS-KGR methods, and show the effectiveness of each module including aliasing relation-assisted module and fusion strategies. Besides, we further analyze the model from various aspects, \textit{i.e.,} transfer ability, computation cost, hyper-parameter analysis, and case study.
\end{itemize}

The paper is structured as follows. Section II provides a concise overview of the related work. Then a detailed introduction of the methodology is presented in Sec.III. After that, Sec. IV illustrated the superiority of the proposed model over sufficient experiments. Finally, Sec. V  concludes the paper.

\section{Related Work}
In this section, we comprehensively introduce the concept and difference of both traditional knowledge graph reasoning task and few-shot knowledge graph reasoning task in Sec. II.A. Furthermore, we divided the few-shot knowledge graph reasoning models into two types, \textit{i.e.}, meta-learning based few-shot KG reasoning methods in Sec. II.B and self-supervised few-shot KG reasoning methods in Sec. II.C. 

\subsection{Knowledge Graph Reasoning}
\paragraph{Traditional Knowledge Graph Reasoning}
The knowledge graph reasoning task is to predict the missing relation between the head entity and tail entity in KG. On account of the inherently symbolic nature of KG, many methods \cite{rw1,rw2} employed logical rules for KG inference in past years. Through achieving high accuracy, these approaches cannot guarantee generalization. To solve the problem, a large amount of representation learning methods have been proposed, aiming to learn distributed representations of entities and relations in KGs. Among them, embedding-based methods \cite{rw3,rw4} usually consider the relation as a translational operation to project the entity embedding into a latent space. Path-based methods \cite{rw5, rw6} is proposed to mine logical information existing in the paths connecting the head entity to the tail entity. GNN-based methods \cite{grail} use graph neural networks to aggregate neighborhood messages, which cover graph structure and entity context information. Though effective, such approaches typically depend on sufficient training instances to learn better embeddings. However, previous works \cite{fsrl} show that the practical KGs consist of a large portion of long-tail relations, which provide insufficient instances to the training process. For example, about 33.7\% of relations in FB15K-237 have no more than 30 instances. Moreover, new relations and entities will be added in KG constantly in practical scenarios, giving KG dynamic properties. 

\paragraph{Few-shot Knowledge Graph Reasoning}
 Compared to traditional KGR methods, FS-KGR is a more daunting task that aims to predict unseen relations in the KG with very limited instances of the new relation \cite{fssurvy}. In this scenario, the model requires learning to generalize from a limited number of instances, meaning that the model understands sufficiently the structure and semantics of the knowledge graph. To the best of our knowledge, most existing FS-KGR approaches typically on the meta-learning framework \cite{fsrl}\cite{gmatching}\cite{metar} while few approaches follow the generative self-supervised learning paradigm \cite{csr}, which have achieved competitive performance.

\subsection{Meta Learning-based Few-Shot KG Reasoning}
Meta-learning is an effective learning paradigm across curating meta-tasks in order to transfer them to a new task \cite{metasurvy}. A large portion of FS-KGR methods employ the meta-learning paradigm to effectively tackle the issue of data scarcity in the target few-shot task \cite{fsrl}\cite{attfmetric}\cite{metar}. These models can be broadly categorized into two types, $i.e.$, metric-based models, and optimization-based models. The former attempts to acquire a universal measure and the corresponding matching functions from a group of training examples and GMatching \cite{gmatching} focus on learning a matching metric that relies on entity embedding and subgraph structure. FSRL\cite{fsrl} proposed a novel method that captures both the different relation types and the different impacts of local neighbors. However, such approaches typically rely on many manual-created meta-training tasks towards specific relations during the training process, leading to poor robustness if the meta-training set is not well-designed. Moreover, creating an excessive number of meta-training sets from the background knowledge graph would eliminate a large proportion of edges, result in sparsity in the graph that hard to learn since the number of relations is limited in real KG, which limits the expressive ability of the model.


\subsection{Self-Supervised Few-Shot KG Reasoning}
In recent years, self-supervised learning has garnered significant attention in various domains like computer vision \cite{cv1,cv2}, natural language processing \cite{nlp1,nlp2}, graph learning \cite{liuyue_DCRN,liuyue_survey,liuyue_SCGC,liuyue_HSAN}, and knowledge graph representation learning \cite{KRACL, SymCL}.


Inspired by their success, some FS-KGR approaches design a powerful self-supervised pretraining objective instead of constructing meta-training tasks, which significantly improves the performance on FS-KGR tasks. Among them, CSR \cite{csr} regarded FS-KGR task as an inductive reasoning task. Concretely, CSR trains an encoder-decoder model in a generative self-supervised learning manner to find and encode the common subgraph structure of each relation. After that, the decoder is used to reconstruct the common subgraph to guarantee the consistency of the subgraph structure before and after reconstruction. Such structure-depended approaches achieve satisfied performance without constructing few-shot training tasks. However, subgraph structure information around the target triplet is insufficient because of the incompleteness of KGs. Moreover, existing SSL-based FS-KGR approaches ignore the logic patterns of data-rich relations that are useful to the task, named aliasing relations, \textit{i.e.}, those relations which are semantically similar to the target relation. Therefore, we novelly proposed an approach in this work for few-shot knowledge graph reasoning by using aliasing relation to assist inference. We demonstrate that the utilization of an aliasing relation can lead to a remarkable enhancement in the few-shot knowledge graph learning task.

\begin{table}  
\centering
\caption{Symbol Notations of SARF}
\renewcommand\arraystretch{1.3}
 \label{notation}
 \begin{tabular}{cc}  
\toprule   
  Symbols & Definition \\  
\midrule   
    $\boldsymbol{G}=\{\boldsymbol{E},\boldsymbol{R},\boldsymbol{T}\}$ & a knowledge graph   \\  
    $\boldsymbol{S_r}$ &  support set of relation $r_i$   \\    
    $K$ & instance number of support set $S_i$ \\
    $\mathcal{M}_i$ & edge mask   \\
    $\rm{C_E}$ & co-occurrence extractor    \\
    $\rm {f}_e(\boldsymbol{G}, \mathcal{M})$ & graph encoder    \\
    $\textbf{E}_g$ & embedding of co-occurrence pattern    \\
    $\rm {f_d}(\boldsymbol{G}, \textbf{E}_g)$ & graph encoder    \\
    $\rm{C_R}$ & co-occurrence reconstructor   \\
    $\textbf{E}_f$ & final embedding of triplet in support set    \\
    $\textbf{E}_t$ & the embedding of relation text \\
    $AR$ & aliasing relations \\
    $\boldsymbol{G}_{AR}$ & AR-subgraph     \\
    $\textbf{E}_{AR}$ & embedding of AR-subgraph    \\
    $m$ & the number of chosen aliasing relations  \\
    $k$ & the sampling number of aliasing triplets  \\
    
  \bottomrule  
\end{tabular}
\end{table}

\section{Methodology}
\begin{figure*}
\centering
\includegraphics[width=500pt]{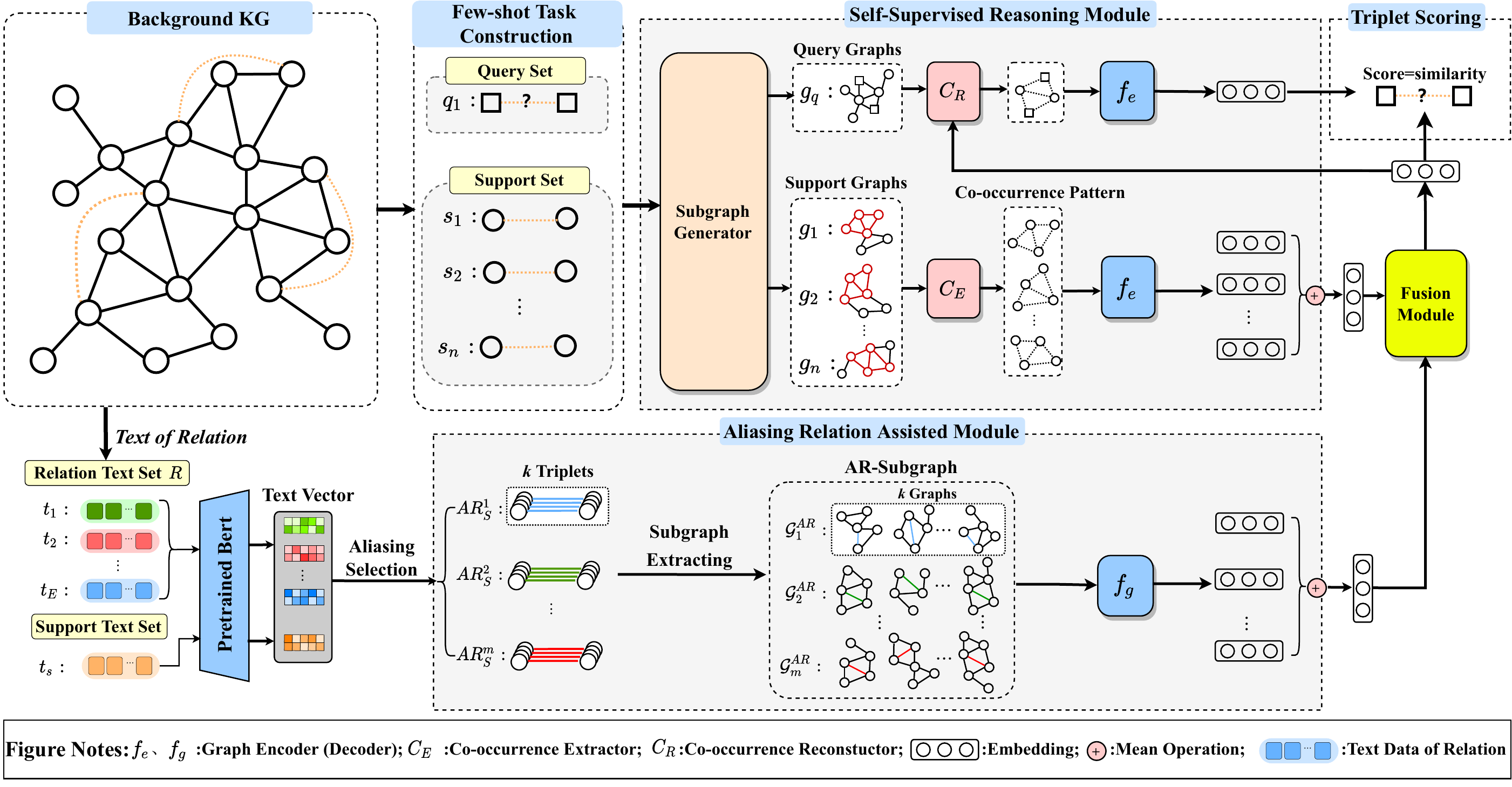}
\caption{The framework of SARF. The SARF includes four main steps: self-supervised reasoning module, aliasing relation assisted module, fusion module, and triplet scoring module. Concretely, in the first step, we construct the few-shot tasks including a support set and a query set corresponding  to the target relation (orange dotted line). In the second step, the co-occurrence patterns are extracted over a co-occurrence extractor $\rm{C_E}$ to represent the target relation. After that, the co-occurrence patterns will be encoded over a graph encoder; In the third step, we select aliasing relations corresponding to the target relation to extract (AR-subgraphs) and encode them into representations. Finally, we fuse the representation of the co-occurrence pattern and AR-subgraphs and test whether there is evidence close enough to the co-occurrence pattern by using a reconstructor $\rm{C_R}$.In general, all the edges shown represent different relations. But in this figure, we specifically highlight the aliasing relations with colors.}
\label{framework}  
\end{figure*}

\vspace{0 em}
In this section, a novel and effective framework termed SARF is proposed for FS-KGR task. The key idea of SARF is to assist reasoning by the meaningful aliasing relations in knowledge graphs. The overview of SARF is shown in figure \ref{framework}. The details of the few-shot task construction, self-supervised reasoning module, aliasing relation-assisted module, and triplet scoring process are introduced in subsection A to D, respectively. 




\subsection{Preliminaries}

In a knowledge graph $\boldsymbol{G}=\{\boldsymbol{E},\boldsymbol{R},\boldsymbol{T}\}$, $\boldsymbol{E}$ denotes the entity set while $\boldsymbol{R}$ denotes the relation set. $\boldsymbol{T}=\{(h_i, r_j, t_k)\}\subseteq\boldsymbol{E}\times\boldsymbol{R}\times\boldsymbol{E}$ denotes the triplet set, where $h_i \in \boldsymbol{E}, r_j \in \boldsymbol{R}, t_k \in \boldsymbol{E}$ is the head entity, relation, and tail entity, respectively. Given a support set $\boldsymbol{S}_r = \{(h_k, t_k)|(h_k, r, t_k)\in\boldsymbol{T},r \in \boldsymbol{R}\}$, the task is to predict triplet $(h,r,t)$ with the relation $r$ missing. 

Here $\left|\boldsymbol{S_r}\right| = K$, which denotes the number of instances in the support set. When $K$ is small, the task is called \emph{K-shot} knowledge graph reasoning. Since there are typically multiple triplets to be predicted during the prediction process, using the support set $\boldsymbol{S_r}$, the collection of all predicted triplets is commonly known as the query set $\boldsymbol{Q_r}={(h, ?, t)}$. In our model, we take advantage of aliasing relations of target relations to assist to in inference. Therefore, an example $s_i = \{(r_i,b_i)|r_i=(h_i, r, t_i) \in \boldsymbol{T}, b_i \in \mathbb{R}^{d \times m} \}$from support sets consist of both triplet and the text sequence of current relation. 
The objective of the FS-KGR task is to develop the ability to predict triplets about an unseen relation $r$ using only limited number of observed triplets associated with $r$. Hence the training process is conducted on well-constructed tasks denoted as $\boldsymbol{T}_{train} = \{\boldsymbol{T_i}\}^{i=1}_M$ where each task $\boldsymbol{T_i} = \{\boldsymbol{S_i},\boldsymbol{Q_i}\}$ comprises a support set and query set. The testing process is subsequently carried out on a set of new tasks denoted as $\boldsymbol{T}_{test}=\{\boldsymbol{T_j}\}^{j=1}_N$, which are similar to $\boldsymbol{T}_{train}$, except that $\boldsymbol{T_j}$ should be about relations not seen in $\boldsymbol{T}_{train}$. However, our model conducts few-shot relation reasoning in a self-supervised manner, which can directly predict the query relation in $\boldsymbol{T}_{test}$ without designing the $\boldsymbol{T}_{train}$. Therefore, the support set and query set mentioned in this paper are all from $\boldsymbol{T}_{test}$. Table \ref{notation} presents a concise overview of the symbol notation used in this section.

\subsection{Self-supervised Reasoning Module}
Inspired by the inductive relation reasoning method \cite{grail}, our insight is that the local subgraph neighborhood of a triplet will contain the latent logical information that indicate the existence of a relation. Specifically, we first seek the approximately shared enclosing subgraph of all triplets in the support set, named co-occurrence pattern, then test whether the pattern is also an enclosing subgraph of candidate query triplet. With this strategy, the model can predict target relation by learning the logical patterns in subgraph structure instead of performing meta-training, which converts the few-shot reasoning task to an inductive reasoning task. 

To allow the model to learn the co-occurrence pattern accurately, we proposed a co-occurrence pattern learning framework, $i.e.$, a self-supervised reasoning module, which includes 3 main components: subgraph generator, co-occurrence extractor, and co-occurrence reconstructor. During the inference time, we first generate an enclosing subgraph of triplets from the support set and query set, then extract the shared subgraph as co-occurrence pattern by leveraging the co-occurrence extractor. 
After that, we encode the co-occurrence pattern and train a co-occurrence reconstructor in a generative self-supervised manner. We hope that the reconstructed co-occurrence pattern should be close to the original one. The details will be illustrated as follows:
\subsubsection{Subgraph Generator}
We presume that the information implying the target relation is present in the pathways linking the two target entities. There are a large number of prior works extract enclosing subgraph around a triplet for relation reasoning \cite{REDGNN, TACT, lm2}, and we choose to utilize the method proposed by GraIL \cite{grail}. Hence, we first sample the $k$-hop (undirected) neighbors around the head and tail entity of each triplet in the support set and query set, denoted as $\mathcal{N}_k(e_h)$ and $\mathcal{N}_k(e_t)$. Then we get the enclosing subgraph by taking the intersection of the neighbor sets, $\mathcal{N}_k(e_h) \cap \mathcal{N}_k(e_t)$ and then eliminate nodes that are isolated or at a distance more than $k$ away from either of the target entities. Specifically, the sparsity of KG determines the value of $k$, which usually be set to 1 or 2. In addition, in cases that the subgraph is insufficiently large, we perform random sampling of the neighbors of $h$ and $t$. The enclosing subgraph induced by support triplets and query triplets are called support graphs and query graphs, respectively.
\subsubsection{Co-occurrence Extractor}
After generating the support graph and query graphs, we try to find the co-occurrence pattern, $i.e.$, the shared subgraph of all support graphs. We design a co-occurrence extractor $\rm{C_E}$ to find the shared subgraph of the support graph. Concretely, we use soft edge mask $\mathcal{M}:[0,1]^\mathbb{R}$ to represent the co-occurrence pattern of these support graphs.
\begin{align}
    \{\mathcal{M}_i\}_{i=1}^K = \rm{C_E}(\{\boldsymbol{G}(h_i, t_i)|(h_i, r', t_i) \in \boldsymbol{S_r'}\})
\end{align}
where $\boldsymbol{G}(h_i,t_i)$ terms the enclosing subgraph induced by head entity $h_i$ and tail entity $t_i$. Specifically, the masks of all support graphs will initialize to be all ones: $\mathcal{M}_i = 1, \forall i\in 1,...,n$. 

Intuitively, $\rm{C_E}$ is designed to compare all support graphs and masks in order to identify the largest shared enclosing subgraph, \textit{i.e.}:
\begin{align}
\rm{argmax}(\sum^{K}_{i=1}{\sum_{E}{\mathcal{M}_i}}),
\end{align}
$$s.t. \forall i,j \in 1...K, \rm{sim}(\rm{f}_e(\boldsymbol{G}_i,\mathcal{M}_i), \rm{f}_e(\boldsymbol{G}_j,\mathcal{M}_j)) > 1 - \varepsilon$$


\begin{algorithm}[tb]
\renewcommand\arraystretch{1.5}
    \caption{Co-occurrence Extractor $\rm{C_E}$}
    \label{alg:algorithm}
    \textbf{Input}: Support graphs set $\{\boldsymbol{G}_i\}_{i=1}^n$\\
    \textbf{Output}: Edge masks $[\mathcal{M}_1,...,\mathcal{M} _n]$ of the support graph
    \begin{algorithmic}[1] 
        \STATE Initialize the masks of $\boldsymbol{G}_i$: $\mathcal{M}_i=1, \forall i \in 1,...,n$
        \FOR{$1 \rightarrow epoch$}
            \FOR{$1\rightarrow p$}
                \FOR{$1\rightarrow q$}
                    \STATE $\mathcal{M}_{pq} = \rm{f}_d(\boldsymbol{G}_p, \rm {f}_e(\boldsymbol{G}_q,\mathcal{M}_q))$
                \ENDFOR
                \STATE $\mathcal{M}_p = \rm{min}_q\mathcal{M}_{pq}$
            \ENDFOR
        \ENDFOR
        \STATE \textbf{return} $[\mathcal{M}_1,...,\mathcal{M}_n]$
    \end{algorithmic}
\end{algorithm}

Here the $\rm{C_E}$ is presented as an iterative procedure to compare all support graph pairs, as shown in algorithm \ref{alg:algorithm}. We first initialize soft edge masks to be all ones for each support graph $\boldsymbol{G}_i$ in step 1. After that, we compared all pairs of $\boldsymbol{G}_p$ to $\boldsymbol{G}_q$ that masked by $\mathcal{M}_q$ to obtains edge mask $\mathcal{M}_{pq}$ in step 2-6. Finally, we takes the minimum of $\mathcal{M}_{pq}$ as $\mathcal{M}_p$ in step 7.

Here, $\rm{f_e}(\boldsymbol{G},\mathcal{M}):\boldsymbol{G} \times \mathbb{R}^{\left| \mathcal{E} \right|} \rightarrow \mathbb{R}^d$ is a graph encoder $\rm{f_e}$ that encode the co-occurrence pattern with corresponding subgraph to embedding, $\rm{f}_d(\boldsymbol{G},\textbf{E}_g):\boldsymbol{G} \times \mathbb{R}^d \rightarrow \mathbb{R}^{\left| \mathcal{E} \right|}$ is a mask decoder that decodes the embedding to masks. After that, we encode the $[\mathcal{M}_1,...,\mathcal{M}_n]$ to over $\rm{f}_e$. To obtain the final embedding $\textbf{E}_g$ of one co-occurrence pattern, we compute the mean of the embedding calculated by $\rm{f}_e$, which encodes subgraph $\boldsymbol{G}$ masked by $\mathcal{M}$ to an embedding $\textbf{g}$.
\begin{align}
    \textbf{g}_i = \rm{f}_e(\boldsymbol{G}_i, \mathcal{M}_i)
\end{align}
\begin{align}
    \textbf{E}_g = \frac{1}{K}\sum_{i=1}^{K}{\textbf{g}_i}
\end{align}
Here $K$ denotes the total number of triplets of the current support set. For the $\rm{f}_e$, we choose PathCon \cite{PathCon} as a GNN encoder, which generates graph embedding with no need for entity embedding:

$$att_v^i = \frac{1}{1+\sum_{e \in \mathcal{N}(v)} {n_e}}\sum_{e \in \mathcal{N}(v)}{\textbf{r}_e^i \cdot n_e}$$     
$$\textbf{r}_v^i = [att_v^i \Vert \mathbbm{1}(v=h)\Vert \mathbbm{1}(v=t)]$$  
$$\textbf{r}^{i+1} = \sigma(([\textbf{r}_v, \textbf{r}_u, \textbf{r}_e]) \cdot \textbf{W} + \textbf{b}^i),u,v \in \mathcal{N}(e)$$  
\begin{align}
    \rm{f}_e(\boldsymbol{G}, \mathcal{M}) = \rm{MaxPool}(att_v^L) \Vert att_h^L \Vert att_t^L,
\end{align}
where $a_v^i$ is attention value of node $v$, $\textbf{r}_e^i$ denotes relation embedding in $i$-th iterations, $L$ denotes the number of layers. Similarly, we use the decoder $\rm{f}_d$ to decode out the mask according to the masked graph $\boldsymbol{G}$ and embedding $\textbf{E}_g$, which is based on the PathCon framework as same as $\rm{f}_e$, except that we summate $\textbf{E}_g$ with the representation of the input edge at first.

\subsubsection{Co-occurrence Reconstructor}

To test the consistency of the co-occurrence pattern between the support graph and query graph, we introduce the co-occurrence reconstructor $\rm{C_R}$, which takes in a query graph $\boldsymbol{G}_q$ and $\textbf{E}_f$ to reconstruct the closest co-occurrence pattern represented by $\textbf{E}_f$.
\begin{align}
    \mathcal{M}_q = \rm{C_R}(\textbf{E}_\textit{f}, \boldsymbol{G}_\textit{q})
\end{align}where $\mathcal{M}_q$ is reconstructed co-occurrence pattern, $\textbf{E}_f$ denotes the final representation which fused $\textbf{E}_g$ and the embedding of aliasing relation $\textbf{E}_{\textit{AR}}$ obtained in section III.C. The score of query triplets is calculated over the similarity function between the embedding $\textbf{E}_f$ and the embedding of the reconstructed co-occurrence pattern.
\begin{align}
    score = \rm{cosine}(\textbf{E}_\textit{f}, \rm{f}_e(\boldsymbol{G_\textit{q}}, \mathcal{M}_\textit{q}))
\end{align}
Specifically, we adopt the consine similarity function to calculate the final score. $\rm{C_R}$ output $\mathcal{M}_q$ if and only if $ \rm{cosine}(\textbf{E}_\textit{f}, \rm{f}_e(\boldsymbol{G}_\textit{q}, \mathcal{M}_\textit{q})) > \varepsilon$, where $\varepsilon$ is a threshold parameter.


\subsection{Aliasing Relation Reasoning}
 Extracting co-occurrence patterns over enclosing subgraphs is an effective way for a few-shot link prediction task. However, the co-occurrence patterns mainly rely on the structure information of the subgraph induced by the current relation, leaving out the logical patterns of data-rich relations which are useful to our task. Concretely, We argue that the relations that are semantically similar to the target relation induce a similar subgraph structure to the target relation that can assist in reasoning, named aliasing relation (AR). To leverage sufficient semantic information of AR, we proposed an aliasing relation assisted method, which mines the logical patterns from aliasing relations of the target relation. The details will be illustrated as follows:

To find aliasing relations, we use a pretrained BERT module to generate the text representation corresponding to the current relation and all relations existing in $\boldsymbol{R}$ respectively.
\begin{align}
    \textbf{E}_T^s = \rm{BERT}(\textit{t}_s)
\end{align}
\begin{align}
    \{\textbf{E}_T^{r_1},...,\textbf{E}_T^{r_{\left|\boldsymbol{R}\right|}}\} = \{\rm{BERT}(\textit{t}_1),...,\rm{BERT}(\textit{t}_{\left|\boldsymbol{R}\right|})\}
\end{align}

Here $t$ denotes the text description of relation, $\textbf{E}_T^{r_i}$ denotes the text embedding of $i$-th relation from relation set $\boldsymbol{R}$. We then select top-$m$ most similar relations $\{AR_1,...,AR_m\}$ by comparing the similarity of $\textbf{E}_T^s$ with $\{\textbf{E}_T^{r_i}\}_{i=1}^{\left|\boldsymbol{R}\right|}$.

For each aliasing relation $AR_i$, we randomly sample $k$ triplets from background KG and then extract the AR-subgraph ${\boldsymbol{G}_{AR_i}} = \{g^1_i,...,g^k_i\}$ over a subgraph extractor that same as Section III.B. To aggregate the information of AR-subgraphs, we compute the mean subgraph embedding that generated by a pretrained graph encoder $\rm{f}_g(g):g \rightarrow \mathbb{R}^d$.
\begin{align}
    \textbf{E}_{AR} = \frac{1}{mk}\sum_{i=1}^{m}{\sum_{j=1}^{k}{\rm{f}_g(\textbf{g}^j_i)}}
\end{align}where our model is robust to the choice of $\rm{f}_g(\cdot)$. In this paper, we choose pretrained GraIL \cite{grail} as the graph encoder $\rm{f}_g$.

Note that we encode the AR-subgraphs directly instead of encoding the co-occurrence pattern of AR-subgraphs extracted by $\rm{C_E}$ to prevent confusion of aliasing relations and target relation. Concretely, the representation of the aliasing relation and target relation will be aligned in a learnable fusion strategy in section D, leading the two representations tend to be consistent with the model training so that the aliasing relations and target relation will be confused during the triplet scoring process. In order to avoid the over-similarity of two representations, we use a coarse method to represent aliasing relations, $i.e.$, the weighted average of all representations of AR-subgraphs. We demonstrate the effectiveness of the aliasing relation representation method in section IV.F.

\subsection{Triplet Scoring and Objective Function}
To fuse the information of aliasing relation and co-occurrence pattern, we introduce a heuristic manner that summate $\textbf{E}_g$ and $\textbf{E}_{AR}$ with fusion rate $\lambda_3$.
\begin{align}
    \textbf{E}_f = \textbf{E}_g + \lambda_3 \textbf{E}_{AR}
\end{align}
To train the whole architecture, we use reconstruction loss $\mathcal{L}_{r}$, $i.e.$, given a KG $\boldsymbol{G}$ and a randomly initialized mask $\mathcal{M}$, the reconstructed mask obtained by applying the co-occurrence reconstructor should match the original mask $\mathcal{M}$, \textit{i.e.}
\begin{align}
    \mathcal{L}_{r} = \rm{CrossEntropy}(\mathcal{M}, \rm{C_R}(\boldsymbol{G},f_e(\boldsymbol{G},\mathcal{M})))
\end{align}

To collaboratively train together with $\rm{f}_e$, a contrastive loss $\mathcal{L}_{c}$ is be introduced:
\begin{align}
    \mathcal{L}_{c} = \rm{max}(\rm{cos\_sim}(\textbf{e}_{pos}, \textbf{e}) - \rm{cos\_sim}(\textbf{e}_{neg},\textbf{e}) + \gamma, 0)
\end{align}where $\textbf{e} = \rm{f}_e(\boldsymbol{G}, \mathcal{M})$ denotes the embedding of graph $\boldsymbol{G}$ masked by $\mathcal{M}$, $\textbf{e}_{pos}$ and $\textbf{e}_{neg}$ denote the embedding of positive and negative graph respectively.
\begin{align}
    \textbf{e}_{pos} = \rm{f}_e(\boldsymbol{G}, \rm{C_R}(\boldsymbol{G}, \rm{f}_e(\boldsymbol{G}, \mathcal{M})))
\end{align}
\begin{align}
    \boldsymbol{G}_{neg} = \rm{f}_e(\boldsymbol{G'}, \rm{C_R}(\boldsymbol{G'}, \rm{f}_e(\boldsymbol{G}, \mathcal{M})))
\end{align}where $\boldsymbol{G}$ and $\boldsymbol{G'}$ are positive and negative graphs respectively, which correspond to different relations. The final loss terms are generally combined with two hyper-parameters $\lambda_1$ and $\lambda_2$ as $\mathcal{L}$.
\begin{align}
    \mathcal{L} = \lambda_1\mathcal{L}_r + \lambda_2\mathcal{L}_c
\end{align}

Note that the performance remains robust to the choice of hyper-parameters. We analyze the influence for performance of three hyper-parameters mentioned above in section IV.E.

As for the summation fusion strategy, we can obtain the optimal performance by tuning fusion rate $\lambda_3$, which is easy to implement. However, the performance and the quality of the final representation heavily depended on the selection of a hyper-parameter, leading to an unstable performance. Besides, the hyper-parameter needs to be re-selected on a new dataset, which is time-consuming. To solve these issues, we introduce a learnable fusion manner that minimizes the MSE loss $\mathcal{L}_{mse}$ between the representation of aliasing relation and co-occurrence pattern. 
\begin{align}
    \mathcal{L}_{\rm{mse}} = \rm{MSE}(\textbf{E}_g, \textbf{E}_{AR})
\end{align}
Intuitively, the learnable fusion manner is an alignment process. The aliasing relation is used to optimize the representation of the co-occurrence pattern without any hyperparameters. The final loss function contains three parts:
\begin{align}
    \mathcal{L} = \lambda_1\mathcal{L}_r + \lambda_2\mathcal{L}_c + \mathcal{L}_{\rm{mse}}
\end{align}
here the optimization goal of SARF is to minimize joint loss.
The optimization goal of SARF (Semantic-Aware Rule Fusion) is to minimize the joint loss

\section{Experiments}
In this section, we illustrate the detailed experimental settings from four aspects, including datasets, evaluation metrics, baselines, and implementation. Subsequently, we conduct comprehensive experiments to verify the superiority of the proposed SARF by answering the main questions as follows.
\begin{itemize}
    \item \textbf{Q1: Superiority.} Does \sysname{} outperforms the state-of-the-art existing few-shot knowledge graph reasoning models?
    \item \textbf{Q2: Effectiveness.} Are the components of the proposed model effective in leveraging the logical pattern of data-rich relations? How do these modules boost performance? 
    \item \textbf{Q3: Transferability.} Can the designed aliasing relation-assisted module be easily and effectively extended to other meta-learning based few-shot knowledge graph reasoning models? 
    \item \textbf{Q4: Sensitivity.} Is the performance of \sysname{} sensitive to  the hyper-parameters, \textit{i.e.,} reconstruction weight $\lambda_1$, contrastive weight $\lambda_2$ and fusion rate $\lambda_3$?
\end{itemize}

To answer the questions mentioned above, we perform a series of extensive experiments. Specifically, answers of \textbf{Q1} to \textbf{Q4} are offered in Sec. IV.B to Sec. IV.E.

\begin{table*}[t]
\renewcommand\arraystretch{1.3}
\fontsize{8}{11}\selectfont 
\caption{Performance comparison of our SARF with other state-of-the-art models on the few-shot knowledge graph link prediction tasks.}
\vspace{-1em}
\resizebox{\linewidth}{!}{
\begin{tabular}{cclcccccccccccccc}
\hline
\multicolumn{3}{c}{\multirow{2}{*}{Methods}}                                      & \multicolumn{4}{c}{NELL}    &  & \multicolumn{4}{c}{FB15K-237} &  & \multicolumn{4}{c}{ConceptNet}  \\ \cline{4-7} \cline{9-12} \cline{14-17} 
\multicolumn{3}{c}{}                                                              & MRR    & Hits\@1    & Hits\@5    & Hits\@10    &  & MRR    & Hits\@1    & Hits\@5    & Hits\@10    &  & MRR    & Hits\@1    & Hits\@5    & Hits\@10    \\ \hline
\multicolumn{17}{c}{Meta-Learning FS-KGR Models}\\\hline
& \multicolumn{2}{c}{GMatching}  & 0.322 & 0.225 & 0.432 & 0.510 &  & \--{} & \--{} & \--{} & \--{} &  & \--{} & \--{} & \--{} & \--{} \\
& \multicolumn{2}{c}{FIRE}  & 0.273 & 0.225 & 0.364 & 0.497 &  & 0.478 & 0.423 & 0.502 & 0.577 &  & \--{} & \--{} & \--{} & \--{} \\
\multicolumn{1}{c}{} & \multicolumn{2}{c}{MetaR}     & 0.471 & 0.322 & 0.647 & 0.763 &  & \textbf{0.805} & \textbf{0.740} & \textbf{0.881} & \textbf{0.937} &  & 0.318 & 0.226 & 0.390 & 0.496 \\
\multicolumn{1}{c}{}

&\multicolumn{2}{c}{FSRL}    & 0.490 & 0.327 & 0.695 & 0.853  &  & {0.684}  & {0.573}  & {0.817}  & {0.912}  &  & 0.577  & {0.469}  & {0.695}  & {0.753}  \\\hline
\multicolumn{17}{c}{Self-Supervised FS-KGR Models}\\\hline

\multicolumn{1}{c}{}  
& \multicolumn{2}{c}{CSR-OPT}  & 0.463 & 0.321 & 0.629 & 0.760 &  & 0.619 & 0.512 & 0.747 & {0.824} &  & 0.559 & {0.450} & 0.692 & 0.736 \\

\multicolumn{1}{c}{}  
& \multicolumn{2}{c}{CSR-GNN}  & 0.560 & 0.435 & 0.703 & 0.821 &  & 0.678 & 0.612 & 0.746 &0.792 &  & \underline{0.615} & \underline{0.518} & 0.729 & 0.757 \\
\multicolumn{1}{c}{}                         & \multicolumn{2}{c}{SARF+Learn} (Ours)     & \textbf{0.627} & \underline{0.493} & \textbf{0.798} & \textbf{0.877} &  & \underline{0.779} & \underline{0.718} & \underline{0.846} & \underline{0.905} &  & 0.613 & 0.511 & \textbf{0.731} & \textbf{0.771} \\ 
\multicolumn{1}{c}{}                         & \multicolumn{2}{c}{SARF+Summat} (Ours)    & \underline{0.626} & \textbf{0.493} & \underline{0.797} & \underline{0.875} &  & 0.753 & 0.688 & 0.814 & 0.884 &  & \textbf{0.624} & \textbf{0.527} & \underline{0.729} & \underline{0.768} \\ \hline
\end{tabular}
}
\label{LP_Result}
\end{table*}

\begin{table}[!t]
\centering
\renewcommand\arraystretch{1.3}
    \fontsize{6}{8}\selectfont  
    \caption{Statistics of all three datasets}
    \vspace{-1em}
    \resizebox{\linewidth}{!}{
        \begin{tabular}{ccccc}
            \hline\cline{1-5} 
    {Dataset} & \multicolumn{1}{c}{$\#\mathcal{E}$}   & \multicolumn{1}{c}{$\#\mathcal{R}$} & \multicolumn{1}{c}{$\#\mathcal{T}$}   & \multicolumn{1}{c}{Tasks}\\ \hline
    NELL             &   68544&    291       & 181109 & 11       \\
    FB15K-237            & 14543 & 200                                    & 268039 & 30    \\ 
    ConceptNet     &   790703 &    14                                         & 2541996   &  2       \\
 \hline
\end{tabular}
}
\label{ashit1}
\end{table}

\subsection{Experiments Setting}
 In this section, we first introduce the experiments setting including three datasets and evaluation metrics, compared baselines and implementation details.
\subsubsection{Datasets and Evaluation Metrics}
We evaluated the proposed SARF on three few-shot benchmarks which generate based on NELL-One \cite{nell}, FB15K-237 \cite{fb} and ConceptNet \cite{cn}. For NELL, we utilized the meta-evaluating and meta-testing splits from NELL-One directly for the evaluation and testing few-shot tasks of NELL. Moreover, we specifically chose relations that had between 50 and 500 triples as few-shot tasks; For FB15K-237 and ConceptNet, we selected a few set ratio of $7:30$ and $1:2$ for target few-shot evaluation and test tasks, respectively, following the approach in \cite{emnlp} and \cite{gmatching}. Table \ref{ashit1} lists the statistics of all three datasets. The model is evaluated based on the standard ranking metrics, \textit{i.e.}, mean reciprocal ranking (MRR) and Hits@$h$, where each test triplet is compared against 50 other candidate negative triplets. We use $h = 1,5,10$. As in the previous study, we report the average results obtained from five times experiments. 

\subsubsection{Baselines}
Our model is compared with two types of few-shot relation reasoning models without curated training tasks, \textit{i.e.}, meta-learning FS-KGR models including GMatching \cite{gmatching}, MetaR \cite{metar}, FIRE \cite{FIRE}, FSRL \cite{fsrl}, and self-supervised FS-KGR models including CSR-OPT and CSR-GNN \cite{csr}. For MetaR and FSRL, we use pretrained entity and relation embedding and randomly sampled meta-tasks for training process instead of utilizing the original meta-split provided in NELL-ONE. For each baseline, we reproduce the optimal results by using their available code and hyperparameters directly.

\subsubsection{Implementation Details}
The total experiments of SARF is in implement on the PyTorch library \cite{pytorch} and conducted on a single NVIDIA GeForce 1080Ti. For the number of few-shot examples $K$, We only take into account $K=3$ for simplicity, even though all the methods in this work can generalize to arbitrary $k$. For the enclosing subgraph generating step, we used $h=2$ hops for NELL, and $h=1$ for both FB15K-237 and ConceptNet. For the graph encoder $\rm{f}_e$ and decoder $\rm{f}_d$ we use 3 layers of PathCon \cite{PathCon} with a hidden dimension of 128. During the aliasing relation selection process, the number of the aliasing relations $m$ and sampled triplets $k$ from background KG is set to 3 and 10 respectively. As for the optimizer for training, we use AdamW with the learning rate of 1e-5. Additionally, the epoch and batch size are configured as 5000 and 8, respectively. Additionally, for the summate fusion process, the loss terms are generally combined as $\mathcal{L}_r + \mathcal{L}_c + \mathcal{L}_{mse}$ with two hyperparameters $\lambda_1$ and $\lambda_2$. On NELL, $\lambda_1 = 1, \lambda_2 = 0.1, and the fusion rate is set to \lambda_3 = 0.01$; On FB15K-237, $\lambda_1 = 1, \lambda_2 = 1, \lambda_3 = 0.01$; On ConceptNet, $\lambda_1 = 2, \lambda_2 = 0.5, \lambda_3 = 10$; Note that, these hyperparameters are robust to the performance in experiments, the more details can see in Section IV.E. 

\subsection{Performance Comparison (RQ1)} 
To answer the Q1, we analysis the superiority and scalability of proposed model, respectively. Concretely, we first compared our SARF with six other state-of-the-art baselines on three datasets, aiming to demonstrate the superiority of proposed model on FS-KGR task. In order to show the scalability of SARF, we compare the data preparation process of SARF with meta-learning based model. After that, we analysis the computational costs, \textit{i.e.}, the inference time of our model and other self-supervised few-shot KG reasoning models.

\subsubsection{Superiority Analysis}
We initially evaluate \sysname{}+Learn and \sysname{}+Summat along with six other models on three datasets without well-designed meta-training tasks. Table \ref{LP_Result} shows that our \sysname{} significantly outperforms other compared baselines for all evaluation metrics.  Taking the best-performing baseline on NELL (CSR-GNN) as an example, the improvement (\%) of \sysname{}+Learn on testing MRR and Hits@1 are 11.96\% and 13.33\%, respectively. With \sysname{}+Summat, the improvement are 11.79\% and 13.33\%, respectively. On FB15k-237, our \sysname{} are second only to MetaR and far superior to other baselines. The reason is that FB15K-237 contains a large number of relations containing only 1 triplet. Compared to MetaR, which trains a meta-relation learner over sufficient training tasks, our \sysname{} trains an expressive encoder and reconstruction instead of designing meta-tasks, which achieves much better empirical performance. 

\begin{figure}[t]
\centering
\includegraphics[width=0.51\textwidth]{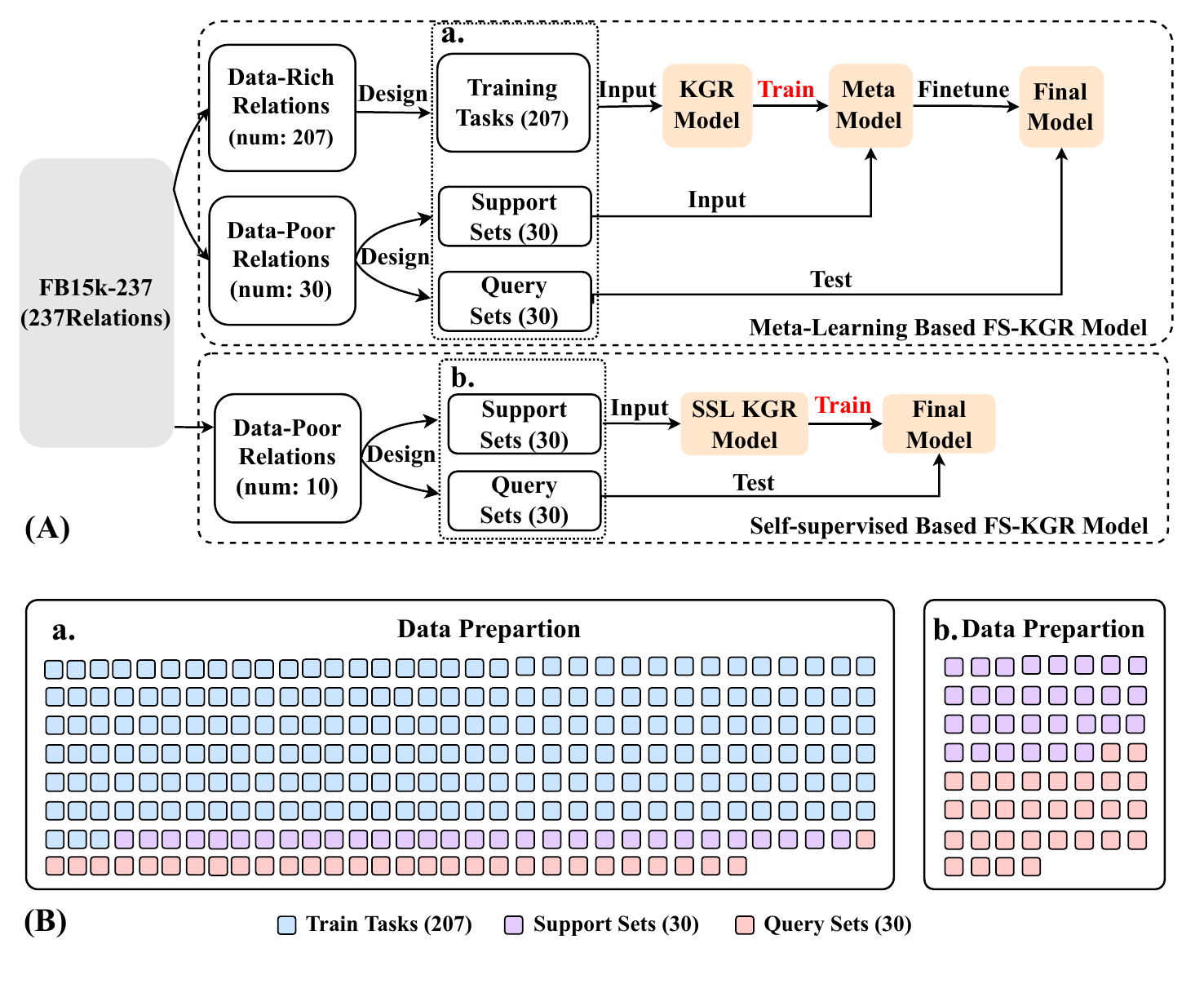}
\caption{The comparison of data preparing process between meta-Learning based FS-KGR methods and SSL-based FS-KGR methods on FB15K-237. The meta-learning based FS-KGR method need to design 207 training tasks, 30 support sets, and 30 query sets manually before training. While SSL-based FS-KGR methods only need 30 support sets and 30 query sets.}
\label{complex}
\end{figure}

\subsubsection{Scalability Analysis} 
It is obvious that digging out more logical patterns from aliasing relations can improve the expressive ability of the few-shot reasoning models, which is proven in previous sections. Moreover, the data preparation process of SARF is simpler than the meta-learning based model. For example on FB15K-237 in figure \ref{complex}(A), the meta-learning based FS-KGR methods need to design 207 training tasks to train a meta model. After that, 30 support sets and query sets are designed manually to fine-tune the meta model and test the final model, respectively. Here each task consists of many instances for specific relations. However, our model only needs to train a co-occurrence pattern encoder and reconstructor over 30 support sets and query sets, saving a lot of time in the data pre-processing process compared to meta-learning based methods. Figure \ref{complex}(B) intuitively shows the number of tasks that need to be designed for meta-learning based FS-KGR methods(a.) and self-supervised based FS-KGR methods(b.), respectively. Apparently, our model is more efficient in the data preparation process than the meta-learning based approach. 

\begin{figure}
    \includegraphics[width=0.47\textwidth]{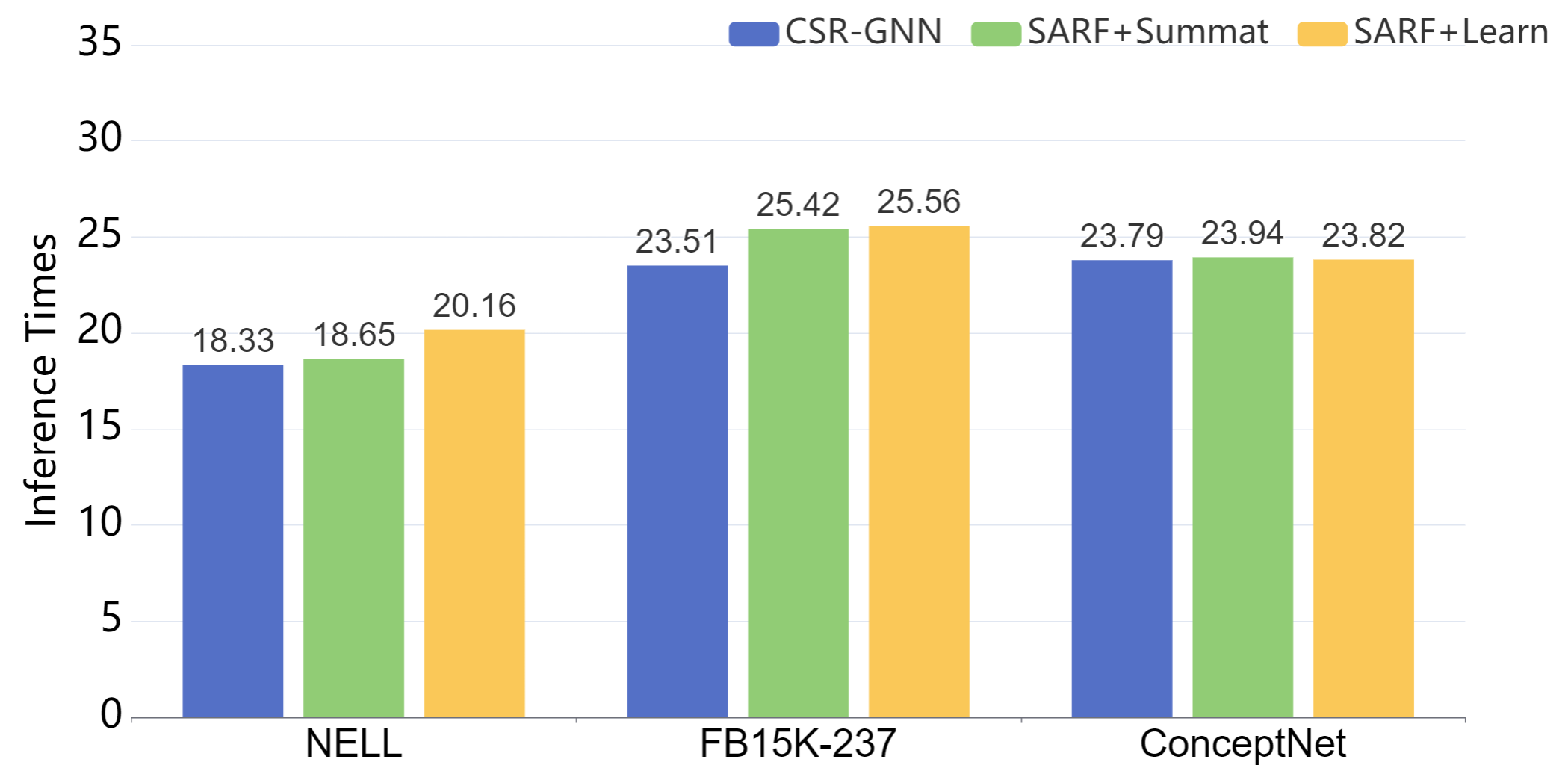}
    \caption{Computational cost analysis of our model and two other baselines in training process. Concretely, We compare the inference time (seconds) of 1 training epoch on three datasets, including NELL, FB15K-237 and ConceptNet.}
    \label{time}
\end{figure}

\begin{table}[!t]
\renewcommand\arraystretch{1.3}
\fontsize{8}{11}\selectfont 
\caption{Ablation study of proposed SARF on three KGs. The model without the aliasing relation assisted module termed SARF w.o. A.R.. In contrast, the full model with summation fusion and learnable fusion strategy are called SARF w. A.R.+ Caoncat. and SARF w. A.R.+ Learn, respectively.}
\resizebox{\linewidth}{!}{
\begin{tabular}{cccccc}
 \hline
{} & Model & MRR   & Hit@1 & Hit@5   & Hit@10\\ \hline
\multirow{3}*{\rotatebox{90}{ NELL }} & SARF w.o. A.R.    &   0.560&    0.435                 & 0.703 & 0.821       \\

& SARF w. A.R. + Summat.      & {0.626{$\uparrow$}} & {0.493{$\uparrow$}}          & {0.797{$\uparrow$}} & {0.875{$\uparrow$}}    \\ 
& SARF w. A.R. + Learn.     &   0.627{$\uparrow$}&    0.493{$\uparrow$}         & 0.798{$\uparrow$}   &  0.875{$\uparrow$}       \\
 \hline

\multirow{3}*{\rotatebox{90}{ FB15K-237 }} & SARF w.o. A.R.             &   0.678&    0.612                 & 0.746 & 0.792       \\
& SARF w. A.R. + Summat.       & {0.753{$\uparrow$}} & {0.688{$\uparrow$}}       & {0.814{$\uparrow$}} & {0.884{$\uparrow$}}    \\ 
& SARF w. A.R. + Learn.     &   0.779{$\uparrow$}&    0.718{$\uparrow$}            & 0.846{$\uparrow$}   &  0.905{$\uparrow$}       \\
\hline

\multirow{3}*{\rotatebox{90}{ ConceptNet }} & SARF w.o. A.R.           &   0.615             &    0.518              & 0.729               & 0.757       \\
& SARF w. A.R. + Summat.     & {0.624{$\uparrow$}} & {0.527{$\uparrow$}}   & {0.729{$\uparrow$}} & {0.768{$\uparrow$}}    \\ 
& SARF w. A.R. + Learn.     &   0.613{$\uparrow$} &    0.511{$\uparrow$}  & 0.731{$\uparrow$}   &  0.771{$\uparrow$}       \\
\hline

\end{tabular}
}
\label{ablation}
\end{table}

Considering the extraction of AR subgraphs will also increase the computational costs, we further measured the inference time of our method and another self-supervised based FS-KGR baseline, \textit{i.e.}, CSR-GNN on three KGs. As shown in figure \ref{time}, we find that compared with the SOTA model CSR-GNN, our SARF achieves comparable inference runtime on NELL and ConceptNet. The inference time of SARF on FB15K-237 is only two seconds slower than CSR-GNN in each epoch though FB15k-237 is denser than NELL and ConceptNet, which is still acceptable. Besides, the learnable fusion strategy increased the number of parameters for training, but it did not significantly increase the inference time. In conclusion, our SRAF has an acceptable time consumption with significant performance improvement.

\begin{table*}[!t]
\renewcommand\arraystretch{1.5}
\fontsize{3}{3.5}\selectfont
\caption{Case study of SARF on real-world toy cases derived from NELL. The tabel shows the improvement of discriminating ability of model during the triplet scoring process. For the triplet (Bugs, ?,Insects), the candidate relations set contains several most likely candidate target relations. AR denotes the aliasing relations assisted module.}
\resizebox{\linewidth}{!}{
\begin{tabular}{cccc}
\hline
Target Triplet                                       & Candidate Relations    & Score w. AR & Score w.o. AR \\ \hline
\multirow{4}{*}{(Bugs, AnimalSuchAsInsects, Insects)} & AnimalSuchAsInsects    & 0.986       & 0.934         \\
& AnimalListTypeOfAnimal & 0.897       & 0.926         \\
& AnimalPreysOn          & 0.891       & 0.914         \\
& HasHusband             & 0.634       & 0.691         \\ \hline

\end{tabular}
}
\label{casestudy}
\end{table*}

 \begin{figure*}
\centering
\includegraphics[width=1 \textwidth]{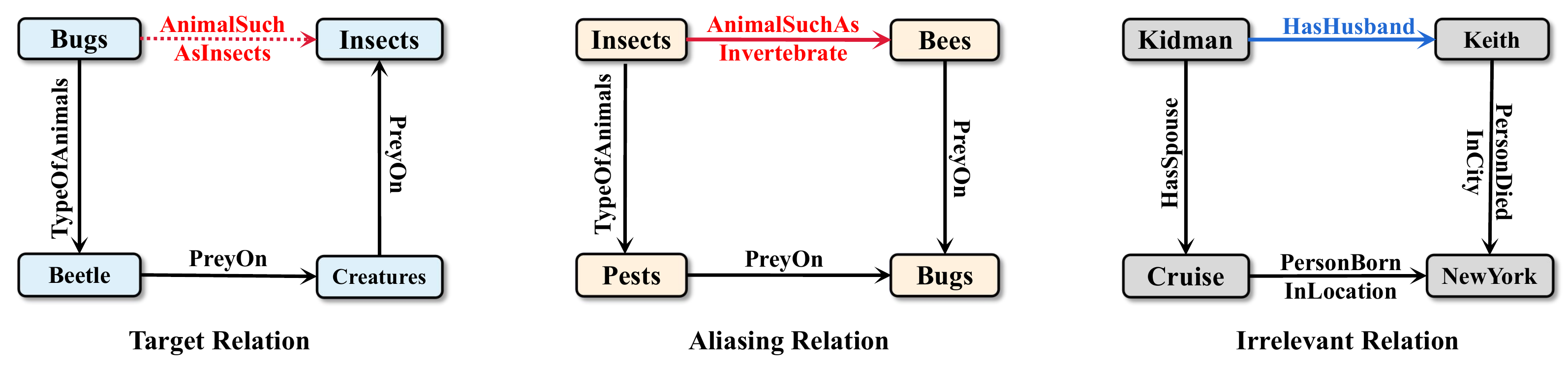}
\caption{Comparison of the subgraph structure induced by the target relation, aliasing relations and irrelevent relation. }
\label{case_study_fig}
\end{figure*}
\subsection{Ablation Study (RQ2)}
The ablation study are performed on multiple datasets to demonstrate the effectiveness and robustness of components of the proposed model in both qualitative and quantitative way, respectively. Specifically, we first gave a intuitive case of aliasing relations on NELL to show the effectiveness of aliasing relations in an explainable manner. To show the contribution for final performance of both the aliasing relation assisted module and two fusion strategies, we conduct experiments on three datasets.



\subsubsection{Case Study for Aliasing Relations}
Aliasing relation is considered to be useful on few-shot KG reasoning task, which contains inherent logical patterns that mentioned in former section. For example, figure \ref{intro} illustrated the aliasing relation selection process briefly with a simple case. In order to qualitatively describe the effectiveness of aliasing relations, we select qualitative examples and visualize a toy case derived from NELL-One. Table \ref{casestudy} shows that the model is more likely to infer the target relation among candidate relations. After introducing the aliasing relation, the inference score gap between the target relation and other candidate relations increases significantly, indicating that the aliasing relation can improve the discriminative ability of our model. As for the target triplet $(Bugs, AnimalSuchAsInsects, Insects)$ in table \ref{casestudy}, one of the corresponding aliasing relation, $i.e.$, $AnimalSuchAsInverbrate$, is highly semantic similar to the target relation $AnimalSuchAsInsects$. Intuitively, the physiological structure and living habits of invertebrates and insects are similar in the real world so that the subgraph structure of the two relations is similar in table \ref{casestudy}. Therefore, the higher score generated by our SARF intuitively illustrates a better discriminative ability.

\subsubsection{Effectiveness of The Designed Modules, \textit{i.e.}, Aliasing Relations Assisted Module and Fusion Module}
To investigate the effectiveness of aliasing relation in a quantitative manner, we removed the aliasing relations assisted module from the origin model. In this setup, the co-occurrence pattern will be reconstructed directly without information fusion. As shown in table \ref{ablation}, the average MRR on NELL, FB15K-237, and ConceptNet up to 11.78\%, 11.06\%, and 1.47\% increase with the aliasing relation assisted module. Meanwhile, to show the effectiveness of different fusion strategies, we compared the performance of SARF with two fusion strategies, respectively. As shown in table \ref{ablation}, the average Hits@10 values are increased by 14.26\% on FB15K-237 with the learnable fusion strategy, while higher than the improvement of 2.65\% brought by leveraging the summation procedure. The experiments result indicates that the learnable fusion strategy is a more effective fusion strategy.

\begin{figure*}
\centering
\includegraphics[width=1 \textwidth]{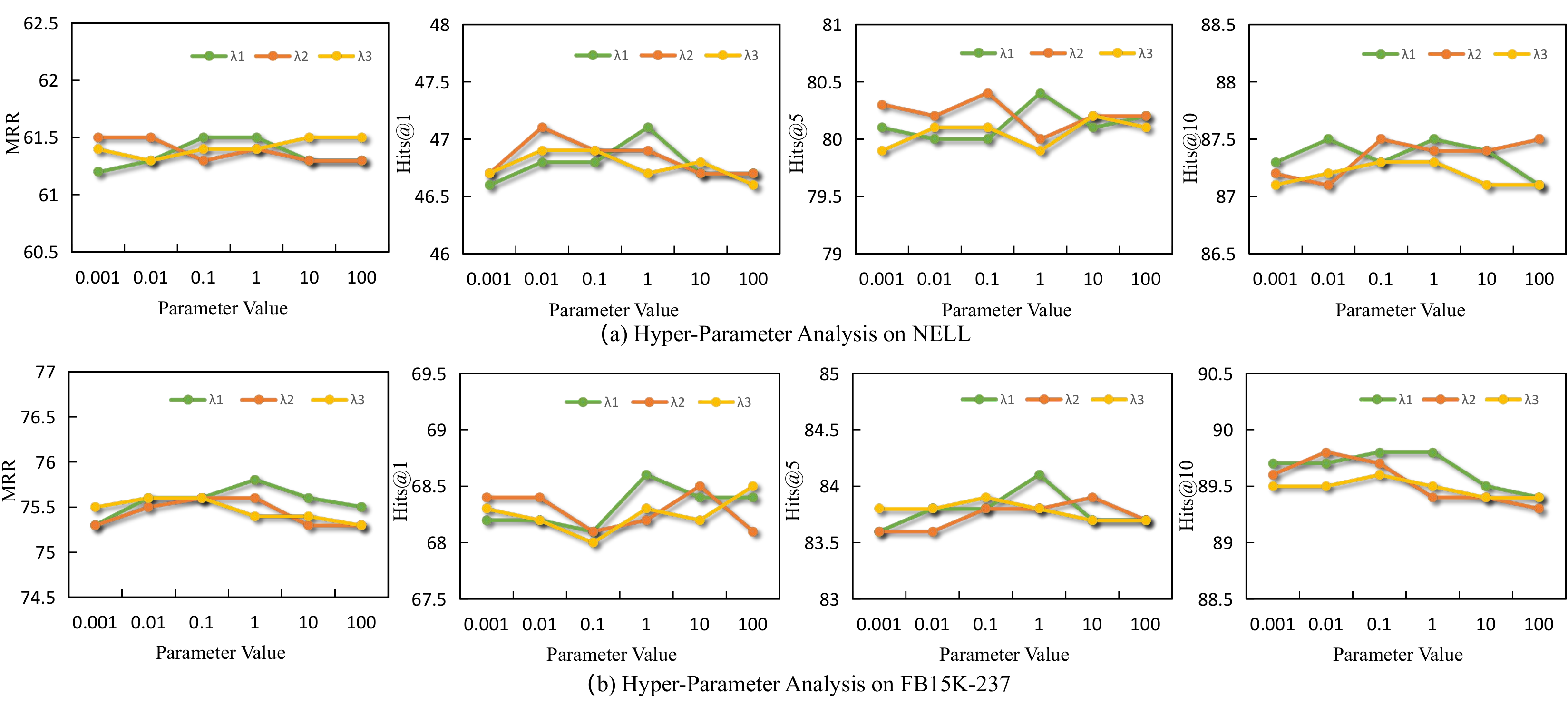}
\caption{Hyper-parameter sensitive analysis of reconstruction weight $\lambda_1$, contrastive weight $\lambda_2$ and fusion rate $\lambda_3$ on NELL and FB15K-237.}
\label{sens}
\end{figure*}

\subsection{Transferability Analysis (RQ3)}
For the purpose of demonstrating the transferability of the proposed model, we extend the main idea of \sysname{}, $i.e$, aliasing relation assisted module, to both of the meta-learning based FS-KGR models, \textit{i.e.}, FSRL, and the self-supervised based FS-KGR model, \textit{i.e.}, CSR. Table \ref{Transf} shows that our \sysname{} can still benefit FSRL and CSR on NELL similar to the conclusions in the previous section. For example to FSRL, there are 7.6\% improvements on MRR and 39.8\% improvements on Hits@1. For CSR, the improvements on MRR and Hits@1 are 11.8\% and 13.3\%, respectively, which demonstrates that the proposed aliasing relation assisted module is a kind of model-agnostic. Therefore, the module can be easily transferred to both meta-learning based and self-supervised based models. Expect the transferability, it also proves the effectiveness of our module from different aspects.

\begin{table}[t]
\renewcommand\arraystretch{1.5}
\fontsize{5}{6}\selectfont 
\caption{Transferability experiments of SARF on the few-shot knowledge graph reasoning tasks.}
\resizebox{\linewidth}{!}{
\begin{tabular}{ccccc}
\hline
\multicolumn{1}{c}{\multirow{2}{*}{Methods}}              & \multicolumn{4}{c}{NELL}      \\ \cline{2-5}  
\multicolumn{1}{c}{}          & MRR    & Hits@1    & Hits@5    & Hits@10        \\ \hline
 CSR & 0.560 & 0.435 & 0.703 & 0.821  \\
 CSR+Learn  & 0.626{$\uparrow$} & 0.493{$\uparrow$} & 0.797{$\uparrow$} & 0.875{$\uparrow$}  \\  \hline
 FSRL & 0.490 & 0.327 & 0.695 & 0.853  \\
 FSRL+Learn  & 0.527{$\uparrow$} & 0.457{$\uparrow$} & 0.691 & 0.860{$\uparrow$}  \\ \hline
\end{tabular}
}
\label{Transf}
\end{table}

\subsection{Sensitive Analysis (RQ4)}
We analyze the influence of three hyper-parameter reconstruction weights $\lambda_1$, contrastive weight $\lambda_2$ and fusion rate $\lambda_3$ for all four evaluation metrics, $i.e.$, MRR, Hits@1, Hits@5, and Hits@10 on two datasets, $i.e.$, NELL and FB15K-237. Concretely, the hyper-parameter reconstruction weight $\lambda_1$, contrastive weight $\lambda_2$ and fusion rate $\lambda_3$ are all selected in \{0.001, 0.01, 0.1, 1, 10, 100\} for both two datasets. To explore the sensitivity of SARF to a particular hyper-parameter, we fix two other hyper-parameters and tune the target one in scope. We observe that the performance for all the metrics only fluctuates within a scope of 0.01 when these hyper-parameters are varying in figure \ref{sens}. The result demonstrates that our SARF is insensitive to $\lambda_1$, $\lambda_2$ and $\lambda_3$. Noted that in order to show the optimal performance, we adopt the best result of different hyper-parameters at all values for all evaluation matrices as the concluding result.

\section{Conclusion}
In this paper, we propose a novel self-supervised based few-shot knowledge graph reasoning model, named \sysname{}, by introducing an aliasing relation assisted module to assist few-shot KG reasoning. Besides, we offered two different fusion strategies, $i.e.$, summation and alignment. As a result, our model achieves better expressive ability due to sufficient information from aliasing relations. Extensive experiments conducted on multiple benchmark datasets for the few-shot link prediction task illustrate that \sysname{} outperforms other compared baselines, achieving the state-of-the-art performance. The result show the effectiveness of aliasing relation-assisted module and two fusion strategies. Besides, we comprehensively analyze various properties of the proposed model from several aspects, \textit{i.e.}, superiority, effectiveness, transferability and sensitivity. In the future, we plan to further optimize the shared subgraph extraction strategy, designing verification mechanism for different subgraph to improve the efficiency of subgraph extraction.

\bibliographystyle{IEEEtran}
\bibliography{sample-base}
\newpage

\vfill

\end{document}